\documentclass{article}

 \usepackage[preprint]{neurips_2026}

\usepackage[most]{tcolorbox}
\usepackage{xcolor}
\usepackage{enumitem}
\usepackage{geometry}
\newcounter{stepnum}

\newtcolorbox{envbox}{
    enhanced,
    breakable,
    colback=white,
    colframe=brown,
    boxrule=0.8pt,
    arc=4pt,
    title={Environment (Step \arabic{stepnum})},
    fonttitle=\bfseries,
    coltitle=white,
    colbacktitle=brown,
    attach boxed title to top left={xshift=4pt,yshift=-2pt},
    boxed title style={
        colback=brown,
        colframe=brown,
        arc=2pt,
        boxrule=0pt
    },
    top=8pt
}

\newtcolorbox{agentbox}{
    enhanced,
    breakable,
    colback=white,
    colframe=blue,
    boxrule=0.8pt,
    arc=4pt,
    title={Agent (Step \arabic{stepnum})},
    fonttitle=\bfseries,
    coltitle=white,
    colbacktitle=blue,
    attach boxed title to top left={xshift=4pt,yshift=-2pt},
    boxed title style={
        colback=blue,
        colframe=blue,
        arc=2pt,
        boxrule=0pt
    },
    top=8pt
}

\usepackage[utf8]{inputenc} 
\usepackage[T1]{fontenc}    
\usepackage{url}            
\usepackage{booktabs}       
\usepackage{amsfonts}       
\usepackage{nicefrac}       
\usepackage{microtype}      
\usepackage{xcolor}         

\usepackage{amsmath,amssymb,amsthm,mathtools}

\usepackage{graphicx}
\usepackage{subcaption}
\usepackage{array}
\usepackage{rotating}
\usepackage[ruled,linesnumbered]{algorithm2e}
\usepackage{float}
\usepackage[table]{xcolor}

\usepackage{multirow}
\usepackage{makecell}   
\usepackage{graphicx}   
\usepackage{wrapfig}
\usepackage{hyperref}       
\usepackage[capitalize]{cleveref}

\crefname{assumption}{Assumption}{Assumptions}
\crefname{lemma}{Lemma}{Lemmas}
\crefname{theorem}{Theorem}{Theorems}
\crefname{definition}{Definition}{Definitions}
\crefname{proposition}{Proposition}{Propositions}

\title{3SPO: State-Score-Supervised Policy Optimization for LLM Agents}

\author{
    Yu Han$^{*1}$, Kailing Li$^{*1}$, Yang Jiao$^2$, Yulin Dai$^1$,\\ 
    \textbf{Yuqian Fu}$^3$, \textbf{Linhai Zhuo}$^4$, \textbf{Tianwen Qian}$^{\dagger1}$\\
    $^1$School of Computer Science and Technology, East China Normal University \\
    $^2$Fudan University, 
    $^3$KAUST, 
    $^4$Fuzhou University \\
    \small{10245102432@stu.ecnu.edu.cn, twqian@cs.ecnu.edu.cn}
    \thanks{$^*$These authors contributed equally to this work.}
    \thanks{$^\dagger$Corresponding author.}
}
\makeatletter
\def\thanks#1{\protected@xdef\@thanks{\@thanks
        \protect\footnotetext{#1}}}
\makeatother
\begin{document}

\maketitle

\begin{abstract}
Training large language models (LLMs) as autonomous agents via reinforcement learning (RL) has enabled frontier models to achieve superhuman performance in long-horizon tasks.
However, existing RL algorithms operate at the trajectory level, performing policy optimization only after collecting complete episode rollouts. This coarse-grained approach faces fundamental challenges in multi-turn agent settings where rewards are sparse, delayed, and credit assignment across individual steps is critical. 
In this work, we propose \textbf{State-Score-Supervised Policy Optimization (3SPO)}, a novel RL algorithm that performs post-step policy optimization with dynamic state score supervision. At each step, 3SPO computes the state score based on historical success rates, supervising step-wise credit assignment, adaptive rollout and post-step policy optimization without requiring value function estimation or additional auxiliary models.
Theoretically, under a per-state bandit abstraction, we show that the proposed score-supervised allocation mechanism achieves logarithmic allocation regret and provide sample-complexity guarantees for action identification, score distinguishability, and filtering stability.
Experiments on ALFWorld and WebShop with Qwen2.5-1.5B/7B-Instruct show that 3SPO consistently outperforms GRPO by $+22.6\%$ on ALFWorld and $+15.6$ points on WebShop, while using comparable resources to achieve $2.4\times$ more state exploration and $1.8\times$ faster convergence.
Code is available at \url{https://github.com/genalyu/3SPO}.
\end{abstract}

\section{Introduction}
\label{sec:intro}

Large language models (LLMs)~\cite{openai2023gpt4,team2023gemini,qwen2025,guo2025deepseekr1} have demonstrated remarkable capabilities as autonomous agents across diverse interactive environments. In embodied task planning~\cite{shridhar2020alfworld}, agents navigate household spaces and complete multi-step goals; in web interaction~\cite{yao2022webshop}, they browse, filter, and purchase items matching complex queries; in code generation~\cite{roziere2023codellama}, they synthesize programs from natural-language specifications. In these multi-turn settings, agents must perceive states, reason over history, and perform long-horizon planning and decision.

\begin{figure}[t]
    \centering
    \includegraphics[width=0.95\textwidth]{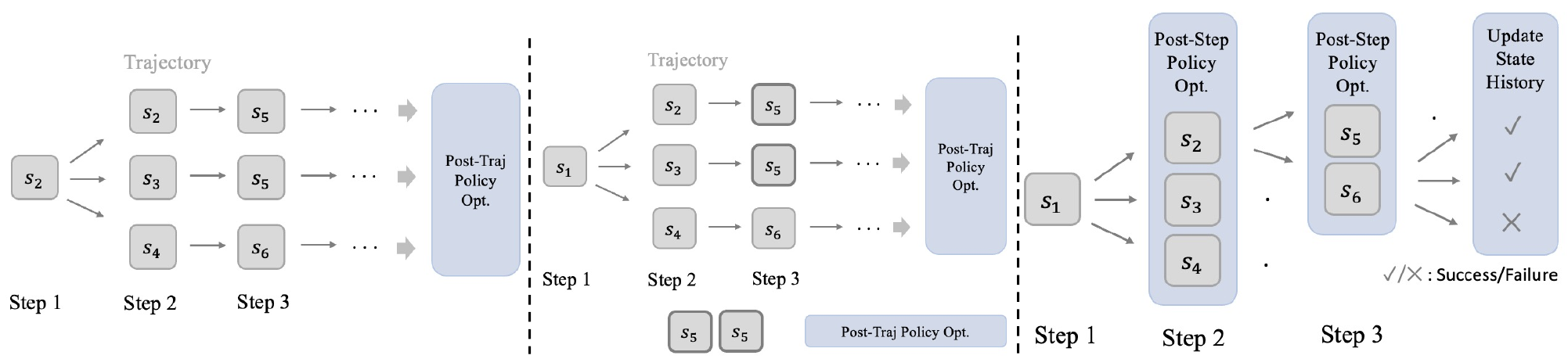}
    \caption{Comparison of multi-turn LLM agent training. \textbf{Left}: Rollout of GRPO is a group of trajectories. \textbf{Middle}: GiGPO introduces state-level grouping but still performs post-trajectory policy optimization. \textbf{Right}: 3SPO computes a continuous state score from historical interaction statistics that simultaneously supervises step-wise credit assignment and adaptive rollout allocation, enabling post-step policy optimization.}

    \label{fig:comparison}
\end{figure}

Group-based reinforcement learning has emerged as a primary approach for training LLM agents through environment interaction. Critic-free methods such as GRPO~\cite{shao2024deepseekmath}, RLOO~\cite{ahmadian2024rloo}, and their variants~\cite{hu2025reinforcepp, yu2025dapo} compute relative advantages within groups of rollouts, avoiding the memory overhead and estimation errors of learned value functions. However, when applied to long-horizon agent tasks, these methods typically operate at the trajectory level: a single episode-level return is assigned to all actions in a rollout, regardless of their individual contribution. This uniform advantage signal obscures fine-grained distinctions across episodes spanning dozens of steps, while sparse terminal rewards provide little guidance on whether intermediate decisions make progress or lead the agent into failure modes. 
Recent attempts address this issue through experience replay~\cite{lu2025arpo}, semi-online RL~\cite{lu2025uis1}, self-refinement loops~\cite{madaan2023selfrefine}, or step-wise credit assignment methods such as GiGPO~\cite{feng2025gigpo} and HGPO~\cite{jiang2025hgpo}. While these approaches recover more informative training signals than trajectory-level baselines, they either require domain-specific engineering or still defer policy updates until complete trajectories are collected. As a result, incorrect intermediate decisions may propagate through subsequent steps before the policy receives any corrective signal, making optimization inefficient and unstable in multi-turn agent environments.

A key observation is that, in multi-turn agent tasks, environment states are not merely transient contexts but reusable decision points whose historical outcomes reveal how well the current policy has mastered the task. States that frequently lead to successful completion are likely mastered, whereas states with low but non-zero success probability often indicate learnable bottlenecks where exploration are most needed. This motivates using state interaction statistics as an online proxy for task completion status and learning potential, enabling step-wise credit assignment and training.

To instantiate this idea, we propose \textbf{State-Score-Supervised Policy Optimization (3SPO)}, a novel RL algorithm that performs step-wise credit assignment, adaptive rollout and post-step policy optimization with dynamic state-score-supervision. The state score measures how unresolved a state remains under the current policy: states with high historical success rates receive lower scores because they are likely mastered, whereas states with low but still recoverable success rates receive higher scores because they expose difficult yet learnable bottlenecks. In this sense, $S(s_t)$ converts sparse trajectory-level outcomes into an online step-level training signal. As illustrated in Fig.~\ref{fig:comparison}, 3SPO simultaneously: (i) governs adaptive rollout allocation, concentrating computation on unresolved states; and (ii) enables post-step policy optimization with a composite reward that balances exploration, credit assignment, and task performance.
Both components arise naturally from different uses of the state score. Specifically, the score difference $S(s_t)-S(s_{t+1})$ provides transition-level credit, the magnitude of $S(s_t)$ determines sampling priority, and the online availability of the score enables policy updates immediately after each step rather than after trajectory completion.

We provide both theoretical and empirical validation of 3SPO. Theoretically, we analyze 3SPO under a per-state bandit abstraction and establish convergence guarantees for the state score mechanism. Empirically, we evaluate 3SPO on ALFWorld and WebShop with Qwen2.5-1.5B/7B-Instruct. Our contributions are as follows:
\begin{itemize}
    \item We introduce a dynamic state score derived from historical interaction statistics, which measures task computation status and learning potential and provides an online step-level signal for agent training. We also provide theoretical guarantees for the state score mechanism.
    \item We propose \textbf{3SPO}, a step-level RL algorithm that uses the state score to supervise reward construction, adaptive rollout allocation, and post-step policy optimization for LLM agents. 
    \item Experiments on ALFWorld and WebShop show that 3SPO consistently outperforms strong prompting and RL baselines, achieving state-of-the-art results on the evaluated benchmarks.
\end{itemize}

\section{Related Work}
\label{sec:related}

\textbf{LLMs as Agents.}
LLMs have been deployed as agents across diverse interactive environments. Prompting-based methods such as ReAct~\cite{yao2023react} and Reflexion~\cite{shinn2023reflexion} operate with frozen parameters, interleaving reasoning traces with action execution and using self-generated feedback for iterative improvement. Tool-augmented LLMs~\cite{qin2024toolllm, shen2024hugginggpt} connect to external APIs and services, enabling information retrieval and service invocation, while memory-augmented architectures~\cite{packer2024memgpt, park2023generative} address long-term context management through hierarchical memory and social simulation.
Despite their breadth of applications, these methods share a fundamental limitation: without parameter updates, the policy cannot adapt its behavior based on accumulated experience, making them brittle in environments requiring systematic exploration or recovery from failure modes not covered by pretraining.

\textbf{Reinforcement Learning for LLMs.}
PPO~\cite{schulman2017ppo} has served as the backbone of RLHF pipelines~\cite{ouyang2022rlhf, bai2022constitutional}, building on trust region theory~\cite{schulman2015trust, kakade2002natural} and GAE~\cite{schulman2016gae}, though its learned value function increases memory overhead and introduces estimation errors. Critic-free group-based methods such as GRPO~\cite{shao2024deepseekmath}, RLOO~\cite{ahmadian2024rloo, kool2019buy}, REINFORCE++~\cite{hu2025reinforcepp}, and DAPO~\cite{yu2025dapo} compute relative advantages within rollout groups, eliminating the need for a value network while maintaining training stability. 
These methods have been primarily demonstrated on single-turn tasks such as mathematical reasoning~\cite{cobbe2021gsm8k, hendrycks2021math, guo2025deepseekr1} and code generation~\cite{roziere2023codellama}. However, extending group-based RL to multi-turn agent settings requires per-step credit assignment, as trajectory-level advantages collapse fine-grained distinctions across long decision sequences.

\textbf{Step-Wise Credit Assignment for LLM Agents.}
Credit assignment in long-horizon tasks with sparse rewards has been addressed through temporal difference learning~\cite{sutton1988td, sutton1998rlbook}, GAE~\cite{schulman2016gae}, count-based exploration~\cite{bellemare2016count, ostrovski2017count}, curiosity-driven methods~\cite{burda2019rnd, pathak2017curiosity, fu2017ex2}, counterfactual baselines~\cite{foerster2018counterfactual, lowe2017multi} and value decomposition~\cite{rashid2018qmix, sunehag2018value}. In LLM agent settings, STEP~\cite{chen2025step} maintains a smoothed success-rate record per state to guide adaptive trajectory resampling and decompose trajectories into step-level training samples. GiGPO~\cite{feng2025gigpo} introduces anchor state grouping for step-level advantage computation, retroactively clustering identical states across completed trajectories. Hierarchical methods~\cite{he2026hierarchy, kulkarni2016hierarchical} structure credit across temporal abstraction levels, while process-supervised rewards~\cite{lightman2023letverify, wang2024processbench} provide step-level feedback at the cost of additional training data and model components. 3SPO's key innovation lies in post-step policy updates rather than deferred trajectory-level optimization.
\section{Method}
\label{sec:method}
\begin{figure}[t]
    \centering
    \includegraphics[width=0.99\textwidth]{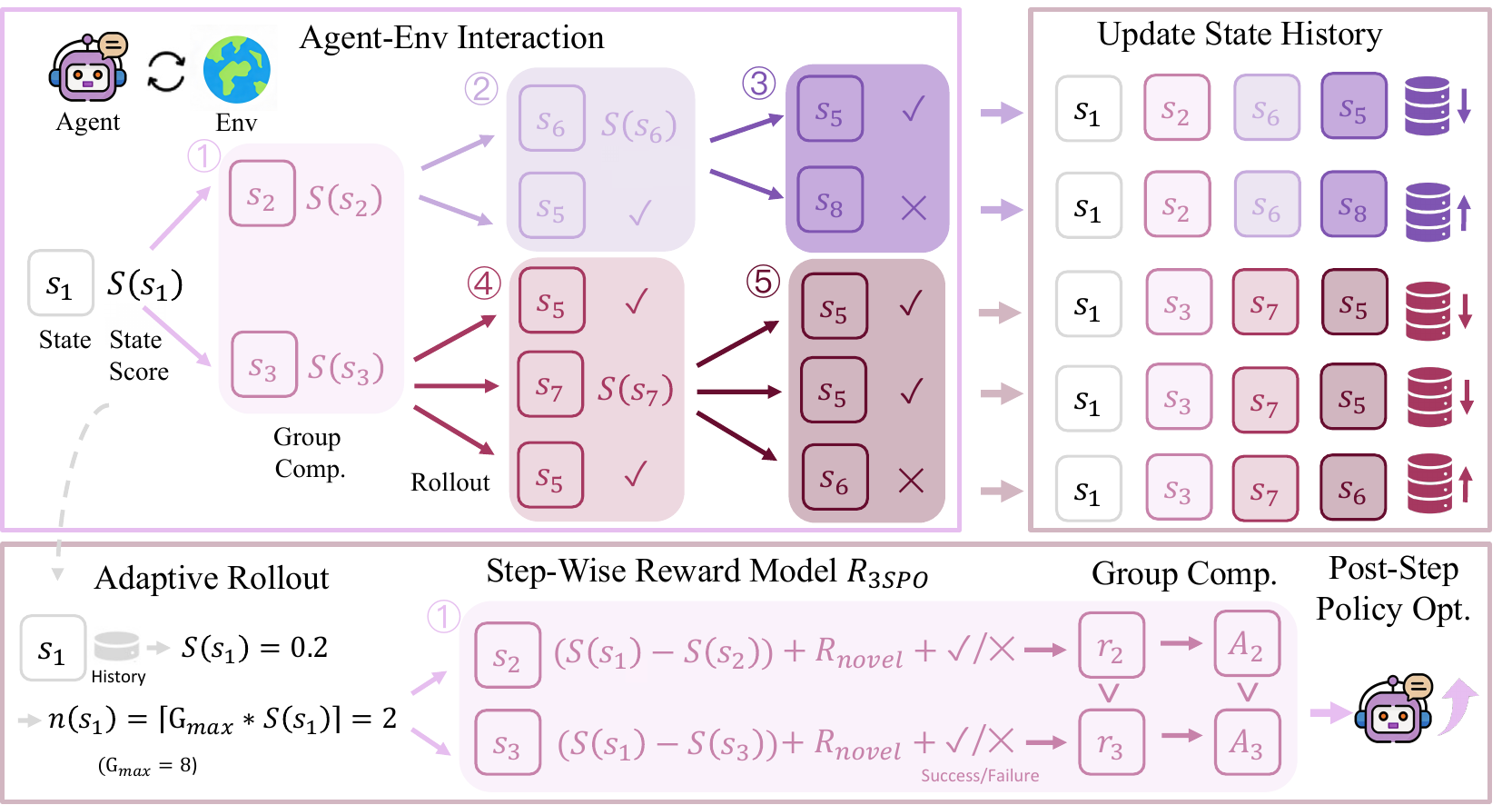}
    \caption{Overview of 3SPO. The agent interacts with the environment to produce state transitions. A dynamic state score $S(s_t)$ computed from historical visit statistics simultaneously supervises: (i) step-wise reward $R_{\text{3SPO}}$, (ii) adaptive rollout allocation $n(s_t)$, and (iii) post-step policy optimization. State history is updated at the end of each trajectory, forming a closed loop learning pipeline.}
    \label{fig:pipeline}
\end{figure}

In this section, we first formulate the multi-turn LLM agent training problem 
(Sec.~\ref{subsec:problem_formulation}) and then introduce 3SPO, a state-score-supervised 
policy optimization framework for long-horizon agentic tasks. As illustrated in 
Fig.~\ref{fig:pipeline}, 3SPO maintains a dynamic state score computed from historical 
interaction statistics and uses it as a unified supervisory signal throughout training. 
The framework consists of three key components: Dynamic State Score Function 
(Sec.~\ref{subsec:state-score func}) quantifies the difficulty and learning potential of each 
state; Step-Wise Reward Model (Sec.~\ref{subsec:reward model}) converts sparse trajectory-level 
outcomes into fine-grained transition-level rewards; Adaptive Rollout and Post-Step Policy 
Optimization (Sec.~\ref{subsec:rollout&optimization}) allocates more rollouts to informative 
states and updates the policy immediately after each step. Finally, we provide theoretical 
insights into the state-score-supervised allocation mechanism in Sec.~\ref{subsec:theory}.

\subsection{Problem Formulation}
\label{subsec:problem_formulation}

The state \(s \in \mathcal{S}\) comprises the current environment observation. At each step $t$, the agent \(\pi_\theta\) parameterized by $\theta$ observes the state \(s_t\) and generates a textual action $a_t \in \mathcal{V}^n$, where $\mathcal{V}$ denotes the token vocabulary and $n$ is the maximum generation length. Upon execution, the environment returns a scalar reward $r_t \in \mathcal{R}$ and the next state $s_{t+1}$. This process repeats until the task is completed or a terminal state is reached, yielding a trajectory $\tau^T  = \{ (s_1, a_1, r_1), (s_2, a_2, r_2), \dots, (s_T, a_T, r_T) \}.$


\subsection{Dynamic State Score Function}
\label{subsec:state-score func}
To quantify the difficulty and learning potential of each interaction state $s_t$ in real time, we propose a dynamic state-score function $S(s_t)$, which is updated smoothly based on the agent's historical interaction statistics on $s_t$:
\begin{equation}
S(s_t) = \exp\left(-\lambda(t) \cdot \frac{N_{\text{success}}(s_t)}{N_{\text{total}}(s_t) + \epsilon}\right) \cdot \mathbf{1}\left\{ N_{\text{fail}}(s_t) < \xi \lor \frac{N_{\text{success}}(s_t)}{N_{\text{total}}(s_t) + \epsilon} > \zeta \right\},
\label{eq:state_score}
\end{equation}
where $N_{\text{total}}(s_t)$ denotes the total number of completed trajectories passing state $s_t$, $N_{\text{success}}(s_t)$ is the number of successful outcomes from $s_t$ and $N_{\text{fail}}(s_t) = N_{\text{total}}(s_t) - N_{\text{success}}(s_t)$ is the number of failures. $\lambda(t) = \alpha \log t$ is a time-varying annealing coefficient that balances exploration at early step and exploitation at later step (the choice of $\alpha$ is detailed in Sec.~\ref{subsec:theory}), $\xi, \zeta > 0$ are threshold hyperparameters that define the valid range of states, and $\epsilon$ is a small constant to avoid division by zero. The state-score function exhibits two key properties: (i) \emph{difficulty-aware learning potential quantification}. The exponential term assigns higher scores to states with lower historical success rates, prioritizing learning on challenging states; (ii) \emph{exploration truncation}. States too difficult to learn are assigned zero score, preventing the algorithm from wasting resources on intractable states that are unlikely to yield meaningful progress.

\subsection{Step-Wise Reward Model}
\label{subsec:reward model}
To resolve the sparse credit assignment problem in long-horizon agent tasks, we propose a step-wise reward model $R_{\text{3SPO}}(s_t, s_{t+1})$, which integrates three complementary reward components:

\begin{equation}
\begin{aligned}
R_{\text{3SPO}}(s_t, s_{t+1}) =\; &\omega(N_{\text{total}}(s_t)) \cdot R_{\text{novel}}(s_{t+1}) \\
&+ \left(0.5 - \omega(N_{\text{total}}(s_t))\right) \cdot (S(s_t) - S(s_{t+1})) \\
&+ 0.5 \cdot R_{\text{success}}(s_{t+1}).
\end{aligned}
\label{eq:reward}
\end{equation}

Specifically: (i) \emph{novelty reward} $R_{\text{novel}}(s_{t+1}) = \mathbf{1}\left\{s_{t+1} \notin\tau^{t} \right\}$ encourages the agent to perform actions that induce observable state changes rather than repetitive or ineffective operations that leave the interface unchanged; (ii) \emph{state score difference reward} $S(s_t) - S(s_{t+1})$ provides credit assignment by rewarding transitions from high-difficulty to low-difficulty states; and (iii) \emph{terminal task reward} $R_{\text{success}}(s_{t+1})$, a score signal indicating task completion. These components remain dynamically balanced via the weight function $\omega(N_{\text{total}}(s_t)) = 0.5 \cdot \exp(-\gamma N_{\text{total}}(s_t))$, which decays exponentially with state visit count (the choice of $\gamma$ is detailed in Sec.~\ref{subsec:theory}). It prioritizes the novelty reward for rarely visited states to encourage exploration, while emphasizes the state score difference reward for frequently visited states to facilitate the exploitation of learned knowledge.

\subsection{Adaptive Rollout and Post-Step Policy Optimization}
\label{subsec:rollout&optimization}

We propose an adaptive rollout rule that dynamically adjusts the number of rollout based on state score to address the inefficiency of uniform sampling. The number of rollout $n(s_t)$ for state $s_t$ is defined as:
\begin{equation}
n(s_t) = \lceil G_{\text{max}} \cdot S(s_t)  \rceil,
\label{eq:rollout}
\end{equation}
where $G_{\text{max}}$ is the maximum number of rollout ($n(s_t) = 1$ proceeds to the next state without policy optimization, and $n(s_t) = 0$ truncates the trajectory immediately). This rollout rule enables adaptive resource allocation: states with higher scores receive more rollout, maximizing the informativeness of each sampling step while reducing waste on mastered states. 

Leveraging the adaptive rollout mechanism together with the state score function and the reward model, we design a post-step policy optimization pipeline without waiting for full trajectory completion. The core optimization objective is:
\begin{equation}
\begin{aligned}
\mathcal{J}_{\text{3SPO}}(\theta) = \mathbb{E}_{\substack{s_t \sim \mathcal{S} \\ \{a^i\}_{i=1}^{n(s_t)} \sim \pi_{\theta_{\text{old}}}}} \Biggl\{ &\frac{1}{n(s_t)} \sum_{i=1}^{n(s_t)} \frac{1}{|a^i|} \sum_{k=1}^{|a^i|} \min \left[ r_{i,k}(\theta) \hat{A}_{i,k}, \text{clip}\bigl(r_{i,k}(\theta), 1-\epsilon, 1+\epsilon\bigr) \hat{A}_{i,k} \right] \\
&- \beta \mathbb{D}_{KL} \left[ \pi_\theta \parallel \pi_{\text{ref}} \right] \Biggr\},
\end{aligned}
\label{eq:objective}
\end{equation}
where $r_{i,k}(\theta) = \frac{\pi_\theta(a^{i,k}|a^{i,<k}, s_t)}{\pi_{\theta_{old}}(a^{i,k}|a^{i,<k}, s_t)}$ is the policy ratio and $\hat{A}_i = \frac{R_{\text{3SPO}i} - \operatorname{mean}(\{R_{\text{3SPO}i}\})}{\operatorname{std}(\{R_{\text{3SPO}i}\})}$ is the group-normalized relative advantage computed using $R_{\text{3SPO}}$. Unlike trajectory-level GRPO variants that aggregate rewards across entire trajectories before updating, 3SPO performs policy optimization immediately after collecting rollouts from the current state $s_t$. 
Instead of committing to a single greedy trajectory, 3SPO employs ranked backtracking DFS. This mechanism not only accelerates state score convergence in the early stages by prioritizing high reward branches but also allows the agent to systematically explore alternative paths whenever a path reaches a termination.
The complete pipeline is summarized in Algorithm~\ref{alg:3spo} (Appendix~\ref{app:algorithm}).

\subsection{Theoretical Insights}
\label{subsec:theory}

We analyze 3SPO under a finite-arm bandit abstraction at each decision state. Each action $a$ at state $s_t$ corresponds to an arm with unknown success probability $p^*(a) \in [0, 1]$, representing the probability of eventual task completion. Given the one-to-one correspondence between each action $a$ and its subsequent state $s_{t+1}$, the state score $S(a) \triangleq S(s_{t+1}) = \exp(-\lambda(t) \hat{p}(a))$ with $\lambda(t) = \alpha \log t$ assigns higher scores to actions with lower empirical success rates $\hat{p}(a)$, supervising rollout allocation toward unresolved states. Let $\mathcal{A} = \{1, \ldots, A\}$ denote the finite action set, $N(a)$ the cumulative samples of action $a$, and $\Delta_a = p^*(a) - p^*(a_w)$ the gap relative to the lowest-success action $a_w = \arg\min_a p^*(a)$. Full proofs of all results are deferred to Appendix~\ref{app:proofs}. Experimental validation is presented in Sec.~\ref{subsec:State_score_analysis}.

\textbf{Concentration of Empirical Estimates.}\label{lem:concentration} Define the ``good event'' at iteration $i$ as $\mathcal{E}_i = \{ \forall a \in \mathcal{A} : |\hat{p}(a) - p^*(a)| \le \sqrt{\frac{\log A}{N(a)}} \}$. The empirical success rate concentrates around its true value under $\mathcal{E}_i$ with probability at least $1 - O(i^{-2})$. By the Borel--Cantelli lemma, $\mathcal{E}_i$ holds almost surely for all but finitely many iterations; its complement $\mathcal{E}_i^c$ (the ``bad event'') has probability decaying as $O(i^{-2})$. This guarantees that historical statistics converge to true success rates .

\textbf{Score Separation.}\label{lem:score_sep} On the concentration event, for any pair of actions $a, b$ with $\Delta_{a,b} = p^*(b) - p^*(a) > 0$ and $N(a), N(b) \ge T_a$, the score difference satisfies $S(a) - S(b) \ge \lambda(t) \exp(-\lambda(t)) \cdot \frac{\Delta_{a,b}}{2}$, where $T_a = \frac{16 \log i}{\Delta_{a,b}^2}$ is the exploration threshold. This proves that the state score reliably distinguishes actions with different success probabilities.

\textbf{Non-worst Action Allocation.}\label{thm:non_worst} For any non-worst action $a \neq a_w$ with $N(a), N(a_w) \ge T_a$, the fraction of rollouts allocated satisfies $\frac{n_i(a)}{\sum_{a'} n_i(a')} \le \exp(-\lambda(t) \cdot \frac{\Delta_a}{2}) = t^{-\frac{\alpha \Delta_a}{2}}$. This ensures that computational resources are preferentially directed toward harder states.

\textbf{Allocation Regret.}\label{thm:regret} We now formally bound the training-time allocation regret of 3SPO's score-supervised rollout allocation. Building on the allocation bound above, the allocation regret over $I$ rounds is:
\begin{equation}
R(I) = \mathbb{E}\!\left[ \sum_{i=1}^I \sum_{a \neq a_w} n_i(a) \cdot \Delta_a \right] = O(\log I).
\end{equation}
This logarithmic bound shows that 3SPO incurs only $O(\log I)$ allocation overhead relative to an oracle unresolved-state sampler under the per-state bandit abstraction. No algorithm can achieve asymptotically smaller regret under the same bandit setting without prior knowledge of the gaps $\Delta_a$. The regret decomposes per action: $R(I) = \sum_{a \neq a_w} \Delta_a \cdot \mathbb{E}[M_I(a)]$, where $M_I(a) = \sum_{i=1}^I n_i(a)$. Each $M_I(a)$ splits into exploration ($N(a) < T_a$) and exploitation ($N(a) \ge T_a$) phases. The exploration phase contributes at most $G_{\max} \cdot T_a$ per action. In the exploitation phase, we invoke the non-worst action allocation bound: the rollout fraction is bounded by $\rho_i(a) = \exp(-\alpha \log t \cdot \Delta_a/2)$ under the good event $\mathcal{E}_i$ defined above. With $\alpha \ge \frac{2 \log I}{\Delta_{\min} \log 2}$ and $t \ge 2$, this gives $\rho_i(a) \le 1/I$. The expectation splits into the good event term ($O(1)$) and the bad event $\mathcal{E}_i^c$ term ($n_i(a) \le G_{\max}$ with probability $O(i^{-2})$, also $O(1)$).

\textbf{Weight Schedule.} The composite reward uses $\omega(N) = 0.5 \exp(-\gamma N)$ to balance novelty and score-difference rewards. At the exploration threshold $T_a = \frac{16 \log I}{\Delta_a^2}$, setting the two weights equal ($\omega = 0.5 - \omega = 0.25$) yields $\gamma = \frac{\log 2}{T_a} = \frac{\Delta_{\min}^2 \log 2}{16 \log I}$. Before $T_a$, $\omega(N) > 0.25$ prioritizes novelty; after $T_a$, the score-difference term dominates.

\section{Experiments}
\label{sec:experiments}

\subsection{Experiment Setup}
\label{ssec:setup}

\textbf{Benchmarks.} We train the LLM agents on ALFWorld~\cite{shridhar2020alfworld} and WebShop~\cite{yao2022webshop}, which are designed to assess the ability of LLM agents to solve long-history tasks. Benchmark details are provided in Appendix~\ref{app:env}.

\textbf{Baselines.} We compare against: (1) Closed-source LLMs: GPT-4o~\cite{openai2023gpt4} and Gemini-2.5-Pro~\cite{team2023gemini}. (2) Prompting agents: ReAct~\cite{yao2023react} and Reflexion~\cite{shinn2023reflexion}. (3) RL training methods: PPO~\cite{schulman2017ppo}, RLOO~\cite{kool2019buy,ahmadian2024rloo}, GRPO~\cite{shao2024deepseekmath}, GiGPO~\cite{feng2025gigpo} and HGPO~\cite{jiang2025hgpo}.

\textbf{Training Details.} We adopt Qwen2.5-1.5B/7B-Instruct~\cite{qwen2025} as our base models. Group-based RL methods use group size $G=8$. For 3SPO, we set $\alpha=50$, $\xi=10$, $\zeta=0.1$ and $G_{\text{max}}=8$. The sensitivity analysis of parameters is presented in Appendix~\ref{app:sensitivity}. Full training settings and hyperparameter details are provided in Appendix~\ref{app:hyperparams}.

\subsection{Experimental Results}

Table~\ref{tab:main_results} shows results across closed models, prompting baselines, and RL training methods. RL training consistently outperforms prompting-only approaches (Qwen2.5, ReAct, Reflexion), with the gap widening on more complex tasks. A pronounced gap exists between trajectory-level methods (PPO, RLOO, GRPO) and step-level methods (GiGPO, HGPO, 3SPO). On both ALFWorld and WebShop, step-level RL-trained methods substantially outperform trajectory-level baselines, confirming that fine-grained credit assignment is crucial for these long-horizon tasks. With Qwen2.5-1.5B-Instruct, 3SPO consistently improves over HGPO by $+2.65\%$ on ALFWorld In-Success and $+3.02\%$ on Out-Success. On WebShop, 3SPO achieves $+5.81\%$ over HGPO in task scores and $+7.92\%$ in task success rates. With Qwen2.5-7B-Instruct, 3SPO remains superior, achieving $96.81\%$ on ALFWorld and $80.45\%$ on WebShop. The gains over HGPO are $+1.37\%/+2.98\%$ on ALFWorld (In/Out-Success) and $+1.94\%$ on WebShop. These suggest that 3SPO's state score supervision scales favorably with model capacity, and enhances model's generalization ability (achieving higher out-success performance compared to other methods). We further observe that 3SPO brings greater gains on the smaller Qwen2.5-1.5B-Instruct model, as its limited agentic capability tends to produce longer and more redundant rollouts that introduce greater bias in advantage estimation. In this case, the state score mechanism becomes more important for refining and stabilizing the credit signal, rollout allocation, and policy optimization. These results suggest that 3SPO is particularly well-suited to challenging long-horizon agentic tasks where rollouts are lengthy and credit signals are noisy.

Fig.~\ref{fig:training_curves}(a) shows success rate training curves on ALFWorld. 3SPO (Blue) converges $1.8\times$ faster than GRPO (Red) and $1.3\times$ faster than HGPO (Pink) and GiGPO (Purple) to reach 60\% success rate. The performance gap between 3SPO and HGPO converges to approximately 6\%, suggesting that state score supervision not only enables faster initialization but also provides sustained benefits throughout training.

\begin{table}[!t]
\centering
\caption{Performance comparison on ALFWorld and WebShop. For ALFWorld, we report the overall success rate ($\uparrow$) for both \emph{in-distribution} (In-Success) and \emph{out-of-distribution} tasks (Out-Success). For WebShop, we report the average task score ($\uparrow$) and the average task success rate ($\uparrow$). Most results are averaged over 3 random seeds during testing. The best results are highlighted in \textbf{bold}, and the second-best results are \underline{underlined}.}
\label{tab:main_results}
\vspace{1em}
\small
\setlength{\tabcolsep}{4pt}
\renewcommand{\arraystretch}{1.0}
\resizebox{0.95\textwidth}{!}{%
\begin{tabular}{@{}lll|cc|cc}
\toprule
Model & Type & Method & \multicolumn{2}{c|}{\textbf{ALFWorld}} & \multicolumn{2}{c}{\textbf{WebShop}} \\
 & & & In-Success & Out-Success & Task Scores & Task Success Rates \\
\midrule
Closed & Prompting & GPT-4o & 48.0 & 46.0 & 31.8 & 23.7 \\
 & Prompting & Gemini-2.5-Pro & 60.3 & 50.5 & 42.5 & 35.9 \\
\midrule
\multirowcell{9}{\small \rotatebox{80}{Qwen2.5-1.5B-Instruct}} & Prompting & Qwen2.5 & 4.1 & -- & 23.1 & 5.2 \\
 & Prompting & ReAct & 12.8 & -- & 40.1 & 11.3 \\
 & Prompting & Reflexion & 21.8 & -- & 55.8 & 21.9 \\
\cmidrule{2-7}
 & RL Training & PPO (with critic) & 54.4$_{\pm3.1}$ & -- & 73.8$_{\pm3.0}$ & 51.5$_{\pm2.9}$ \\
 & RL Training & RLOO & 69.7$_{\pm2.5}$ & 68.7$_{\pm10.7}$ & 73.9$_{\pm5.6}$ & 52.1$_{\pm6.7}$ \\
 & RL Training & GRPO & 72.8$_{\pm3.6}$ & 70.1$_{\pm2.5}$ & 75.8$_{\pm3.5}$ & 56.8$_{\pm3.8}$ \\
\cmidrule{2-7}
 & RL Training & GiGPO & 90.16$_{\pm0.22}$ & 84.76$_{\pm2.83}$ & 84.95$_{\pm2.57}$ & 66.53$_{\pm1.92}$ \\
 & RL Training & HGPO & \underline{92.77}$_{\pm1.08}$ & \underline{90.16}$_{\pm0.78}$ & \underline{85.56}$_{\pm2.86}$ & \underline{71.54}$_{\pm4.00}$ \\
 & RL Training & \textbf{3SPO} & \textbf{95.42}$_{\pm0.58}$ & \textbf{93.18}$_{\pm1.21}$ & \textbf{91.37}$_{\pm0.82}$ & \textbf{79.46}$_{\pm1.73}$ \\
\midrule
\multirowcell{9}{\small \rotatebox{80}{Qwen2.5-7B-Instruct}} & Prompting & Qwen2.5 & 14.8 & -- & 26.4 & 7.8 \\
 & Prompting & ReAct & 31.2 & -- & 46.2 & 19.5 \\
 & Prompting & Reflexion & 42.7 & -- & 58.1 & 28.8 \\
\cmidrule{2-7}
 & RL Training & PPO (with critic) & 77.08$_{\pm1.12}$ & 76.23$_{\pm1.46}$ & 81.4$_{\pm3.1}$ & 68.7$_{\pm5.1}$ \\
 & RL Training & RLOO & 77.86$_{\pm0.03}$ & 73.95$_{\pm0.05}$ & 80.3$_{\pm3.2}$ & 65.7$_{\pm4.0}$ \\
 & RL Training & GRPO & 78.64$_{\pm0.73}$ & 76.82$_{\pm1.47}$ & 79.3$_{\pm2.8}$ & 66.1$_{\pm3.7}$ \\
\cmidrule{2-7}
 & RL Training & GiGPO & 93.29$_{\pm0.40}$ & 92.18$_{\pm0.39}$ & 88.93$_{\pm1.49}$ & 77.60$_{\pm1.68}$ \\
 & RL Training & HGPO & \underline{95.44}$_{\pm0.62}$ & \underline{92.05}$_{\pm0.22}$ & \underline{88.96}$_{\pm1.04}$ & \underline{78.51}$_{\pm1.40}$ \\
 & RL Training & \textbf{3SPO} & \textbf{96.81}$_{\pm0.35}$ & \textbf{95.93}$_{\pm0.67}$ & \textbf{90.28}$_{\pm0.53}$ & \textbf{80.45}$_{\pm0.89}$ \\
\bottomrule
\end{tabular}%
}
\end{table}

\begin{figure*}[tb]
\centering
\setlength{\tabcolsep}{2pt} 

\begin{minipage}[t]{0.24\textwidth}
  \centering
  \includegraphics[width=\linewidth, height=2.2cm]{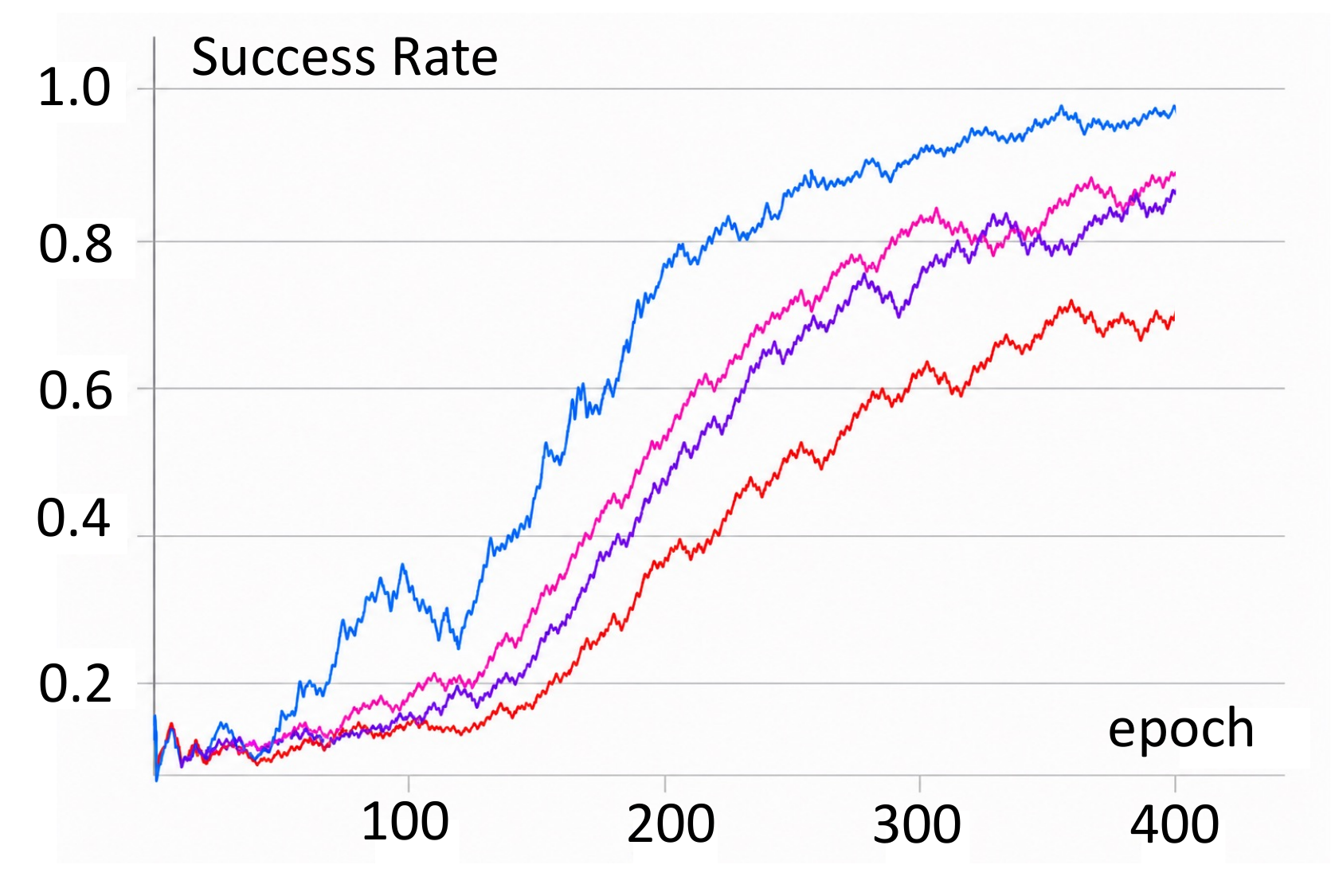}
  \vspace{-1.5em}
  \caption*{\footnotesize (a)}
\end{minipage}
\hfill
\begin{minipage}[t]{0.24\textwidth}
  \centering
  \includegraphics[width=\linewidth, height=2.2cm]{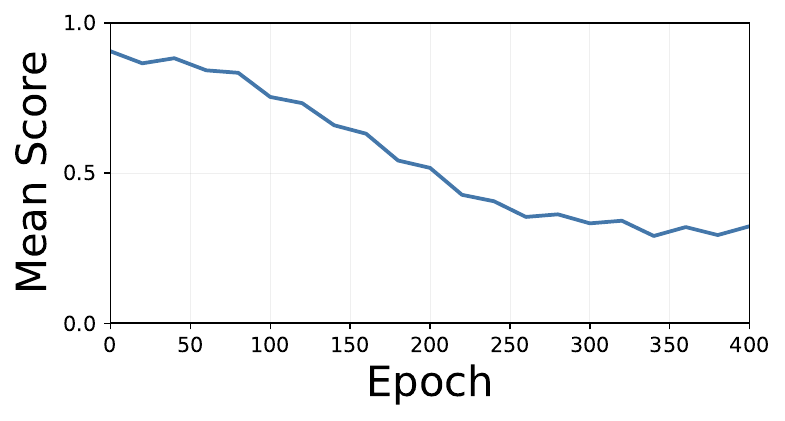}
  \vspace{-1.5em}
  \caption*{\footnotesize (b)}
\end{minipage}
\hfill
\begin{minipage}[t]{0.24\textwidth}
  \centering
  \includegraphics[width=\linewidth, height=2.2cm]{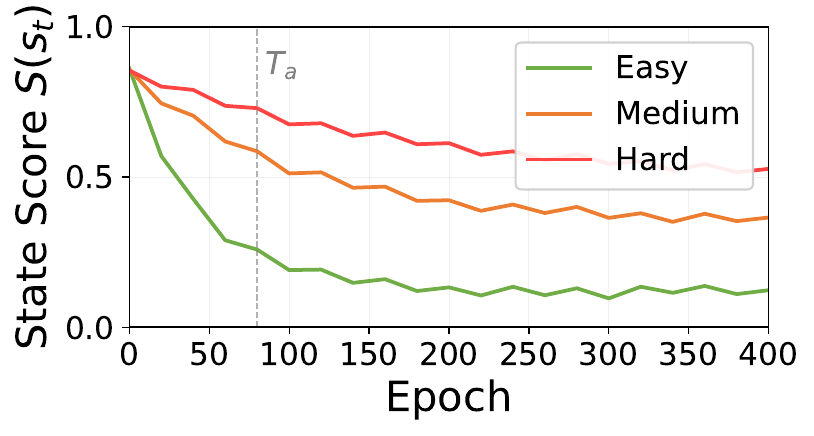}
  \vspace{-1.5em}
  \caption*{\footnotesize (c)}
\end{minipage}
\hfill
\begin{minipage}[t]{0.24\textwidth}
  \centering
  \includegraphics[width=\linewidth, height=2.2cm]{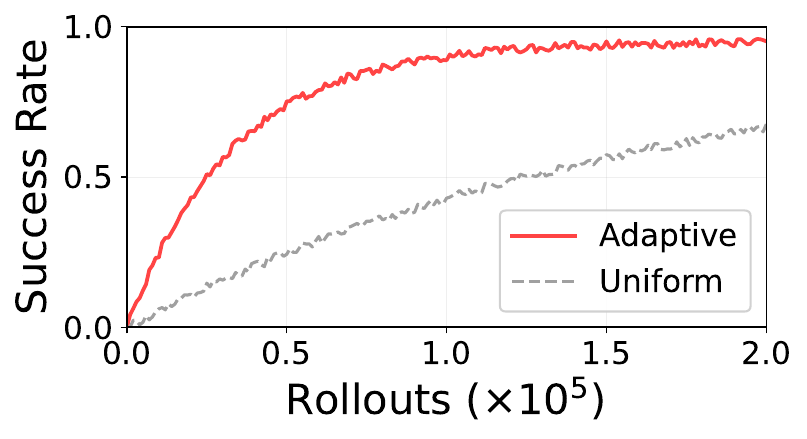}
  \vspace{-1.5em}
  \caption*{\footnotesize (d)}
\end{minipage}

\caption{Training and state score dynamics on ALFWorld (Qwen2.5-1.5B). (a) 3SPO (Blue) converges faster than GRPO (Red), GiGPO (Purple) and HGPO (Pink). (b) Mean state score decays monotonically as the policy improves. (c) Score distributions shift from high to low as states are mastered across easy/medium/hard tasks. (d) Adaptive rollout converges to $>90\%$ success within $2.0\times 10^5$ rollouts, while uniform rollout remains at $\sim\!66\%$ and still rising.}
\vspace{-1.5em}
\label{fig:training_curves}
\end{figure*}

\vspace{-1em}

\subsection{State Score Analysis}
\label{subsec:State_score_analysis}

The state $s_t$ is defined as the complete textual observation at step $t$. Taking ALFWorld as an example, $s_t$ includes the current room description, visible objects, and inventory status.
Fig.~\ref{fig:training_curves} (b--d) examines how the state score mechanism evolves during training:
(b) The mean state score remains high during the initial 70 epochs of exploration, then decays monotonically from $\approx 0.92$ to $\approx 0.35$ by epoch 400. The early plateau reflects the agent discovering new states faster than it masters them; once exploration stabilizes, the score declines as previously ambiguous states are resolved.
(c) We categorize some ALFWorld tasks into easy (Pick \& Place), medium (Examine in Light), and hard (Pick Two \& Place). Easy tasks converge to low scores rapidly, indicating mastery; medium tasks exhibit a slower decay; hard tasks remain elevated throughout training, demonstrating that the score automatically prioritizes harder states.
(d) Adaptive 3SPO reaches $>80\%$ success within $1.0\times 10^5$ rollouts and converges to $>90\%$, whereas uniform rollout ($n(s_t) = G_{\max}$) only reaches $\sim\!66\%$ at $2.0\times 10^5$ rollouts with no sign of convergence. This demonstrates that allocating more rollouts to unresolved states yields a substantially more sample-efficient learning trajectory.




\begin{figure*}[tbp]
\centering
\begin{minipage}[t]{0.53\linewidth}
\centering
\captionsetup{font=footnotesize}
\captionof{table}{Ablation study on ALFWorld and WebShop.}
\label{tab:ablation}
\footnotesize
\setlength{\tabcolsep}{2pt}
\setlength{\extrarowheight}{8pt}
\begin{tabular}{l|cc|cc}
\toprule
Ablation & \multicolumn{2}{c|}{ALFWorld} & \multicolumn{2}{c}{WebShop} \\
 & In-Suc. & Out-Suc. & Task Sco. & Task Suc. \\
\midrule
\rowcolor{gray!15} \textbf{3SPO} & 95.42$_{\pm0.6}$ & 93.18$_{\pm1.2}$ & 91.37$_{\pm0.8}$ & 79.46$_{\pm1.7}$ \\
\midrule
w/o $R_{\text{novel}}$ & 88.42$_{\pm1.7}$ & 85.70$_{\pm2.1}$ & 85.12$_{\pm2.2}$ & 72.06$_{\pm2.2}$ \\
w/o $\Delta S(s_t)$ & 87.34$_{\pm2.1}$ & 83.58$_{\pm2.5}$ & 82.43$_{\pm2.6}$ & 72.46$_{\pm2.6}$ \\
w/o Ada.~$\omega(N)$ & 92.94$_{\pm1.4}$ & 91.02$_{\pm1.8}$ & 90.20$_{\pm1.9}$ & 77.82$_{\pm1.0}$ \\
\midrule
w/o Ada.~$n(s_t)$ & 95.88$_{\pm1.3}$ & 93.42$_{\pm1.6}$ & 91.56$_{\pm1.7}$ & 79.82$_{\pm1.9}$ \\
w/o Backtrack & 88.98$_{\pm1.4}$ & 87.00$_{\pm1.9}$ & 85.32$_{\pm1.9}$ & 74.18$_{\pm2.0}$ \\
\bottomrule
\end{tabular}
\end{minipage}
\begin{minipage}[t]{0.44\linewidth}
\centering
\captionsetup{font=scriptsize}
\captionof{table}{Computational efficiency on ALFWorld.}
\label{tab:efficiency}
\tiny
\setlength{\tabcolsep}{2.6pt}
\begin{tabular}{lccc}
\toprule
\textbf{Method} & \textbf{Rollouts} & \textbf{Unique States} & \textbf{Ratio} \\
\midrule
GRPO &  3.4 $\times 10^5$ & 4,020 & 1.2$\times 10^{-2}$ \\
HGPO & 3.4 $\times 10^5$ & 3,890 & 1.1$\times 10^{-2}$\\
GiGPO &  3.4 $\times 10^5$ & 4,150 & 1.2$\times 10^{-2}$\\
\midrule
\rowcolor{gray!15} \textbf{3SPO} & 4.5 $\times 10^5$ & 9,720 & \textbf{2.2$\times 10^{-2}$} \\
w/o Ada.~$n$  & 3.3 $\times 10^6$ & 10284 & 3.1$\times 10^{-3}$ \\
\bottomrule
\end{tabular}
\includegraphics[width=0.8\linewidth]{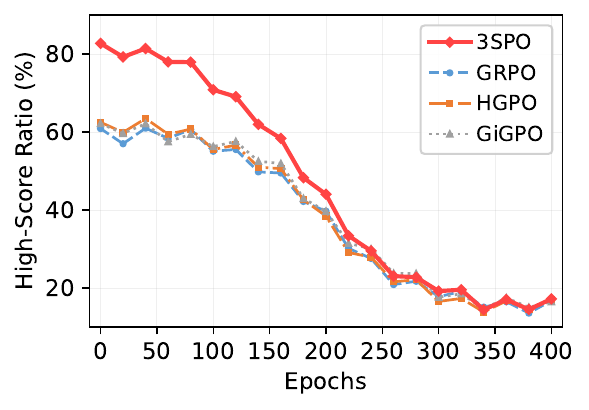}
\vspace{-1.5em}
\captionof{figure}{High-score ratio during training.}
\label{fig:high_score_comparison}
\end{minipage}
\end{figure*}

\subsection{Ablation Study}
In this section, we conduct an ablation study to evaluate the effectiveness of each component in 3SPO.
As shown in Table~\ref{tab:ablation} and~\ref{tab:efficiency}, ``w/o'' denotes removing a specific component.

\textbf{Reward Components.} First, ``w/o $R_{\text{novel}}$'' removes the novelty reward, degrading performance by $7.0\%$ on ALFWorld and $6.8\%$ on WebShop. This validates that encouraging observable state changes prevents repetitive loops. Second, ``w/o $\Delta S(s_t)$'' removes the state score difference reward, causing the largest drop ($-8.1\%$ on ALFWorld, $-9.0\%$ on WebShop). This confirms that credit assignment through state score transitions is the dominant factor in 3SPO's effectiveness. Finally, ``w/o Ada.~$\omega(N)$'' replaces the adaptive weight with a fixed $\omega=0.5$. Performance drops $-2.5\%$ on ALFWorld and $-1.6\%$ on WebShop, suggesting that the adaptive schedule progressively shifts from novelty-driven exploration to score-based exploitation as states are visited more.

\textbf{Adaptive Rollout.} ``w/o Ada.~$n(s_t)$'' replaces adaptive rollout with uniform allocation $n(s_t)=G_{\max}$. Performance is slightly higher by $+0.5\%$ on ALFWorld and $+0.2\%$ on WebShop, but at $7.3\times$ the computational cost and with an extremely low state exploration ratio of only $3.1 \times 10^{-3}$, which is nearly $7\times$ lower than that of the full 3SPO ($2.2 \times 10^{-2}$). The adaptive mechanism allocates computation to states that most need learning signals while avoiding waste on easy states.

\textbf{Ranked Backtracking DFS.} ``w/o Backtrack'' disables the backtracking mechanism, dropping performance by $-6.4\%$ on ALFWorld and $-6.0\%$ on WebShop. This confirms that the ranked backtracking strategy is important for thorough state coverage, without which the agent commits to locally optimal early actions and cannot recover.

\subsection{Computational Efficiency}

Table~\ref{tab:efficiency} compares computational efficiency on state coverage. A state is counted as unique if it has not been encountered in any previous rollout across the entire training process, and this exact-string matching provides a reliable measure of state space coverage. 3SPO visits $\sim\!2.4\times$ more unique states than other methods. Based on the quantitative ratio of unique states to total rollouts, 3SPO achieves an exploration ratio of $2.2 \times 10^{-2}$, nearly doubling that of GRPO, HGPO and GiGPO (around $1.1$--$1.2 \times 10^{-2}$). The backtracking mechanism continuously exposes the agent to previously unvisited states, and adaptive rollout concentrates exploration on unresolved regions.

Fig.~\ref{fig:high_score_comparison} compares the high-score ratio ($S(s_t)>0.5$) trajectory of 3SPO against other RL methods. 3SPO maintains a substantially higher proportion of high-score states throughout training ($\sim\!75\%$ at early epochs vs.\ $\sim\!55$--$60\%$ for GRPO/GiGPO/HGPO). GRPO, GiGPO, and HGPO exhibit similar high-score ratio levels since none of them employ an adaptive rollout mechanism to concentrate sampling on unresolved states. In contrast, 3SPO's adaptive rollout $n(s_t)$ allocates more rollouts to states with higher scores, ensuring that unresolved states receive more learning signals per training step. This elevated high-score ratio in 3SPO reflects a more data-efficient sampling strategy: the agent repeatedly explores unresolved states rather than distributing its budget uniformly.

\subsection{Case Study}
\label{subsec:case_study}
\begin{wrapfigure}{r}{0.55\textwidth}

    \vspace{-1em}
    \centering
    \includegraphics[width=\linewidth]{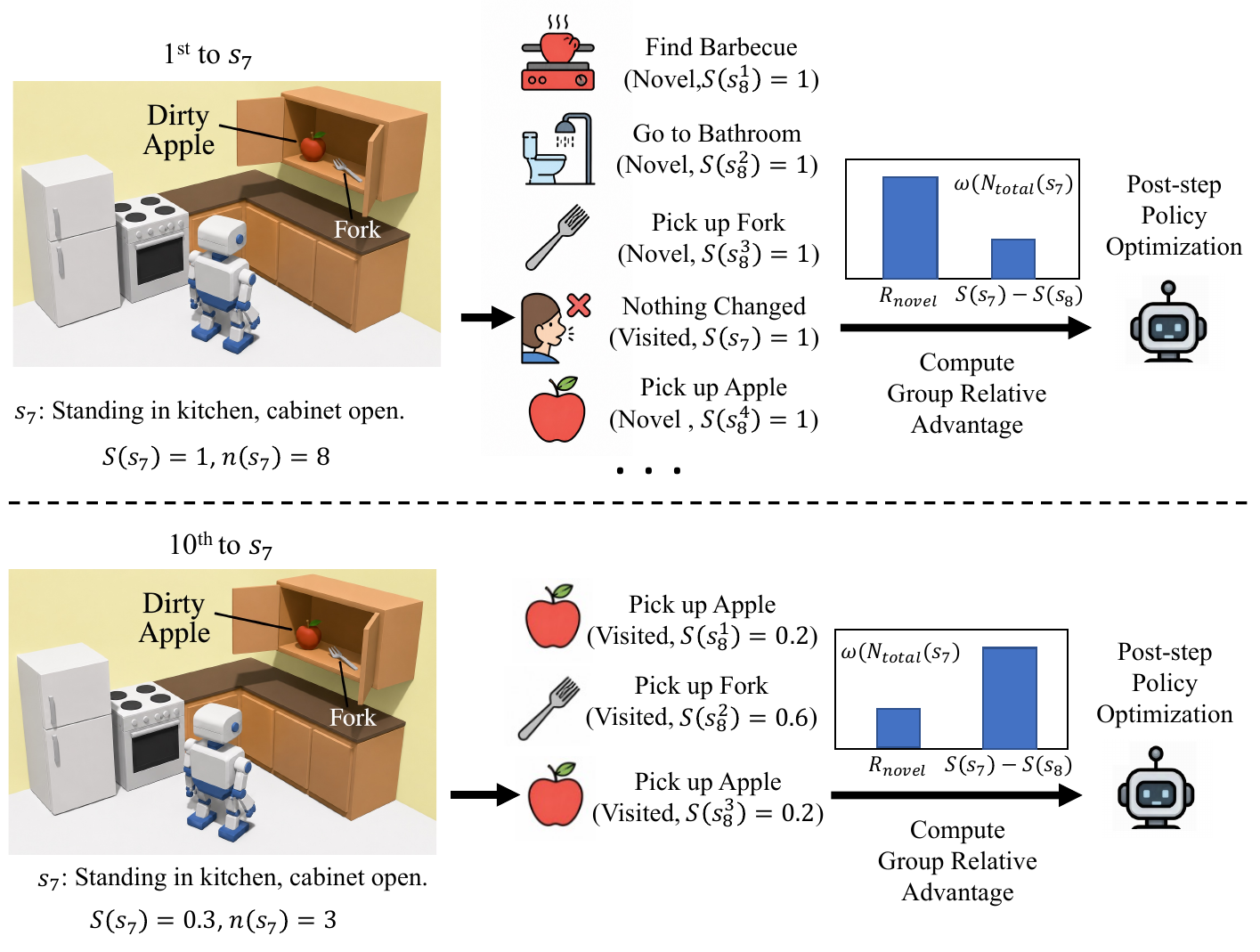}
    \caption{Case study of early and late visits to the same state \(s_7\).}  
    \vspace{-1.7em}
    \label{fig:case_study}
\end{wrapfigure}

As illustrated in Fig.~\ref{fig:case_study}, consider a ``Pick \& Place'' task where the agent must navigate between rooms and manipulate multiple objects. When the agent first visits a state $s_7$ (e.g., standing in the kitchen after opening a cabinet and finding a dirty apple and a fork), the state score $S(s_7) = 1$ due to zero prior visits. The adaptive rollout rule allocates $n(s_7) = \lceil G_{\max} \cdot S(s_7) \rceil = 8$ rollouts from this state.
Each rollout produces next state $s_8^{(i)}$: some proceed to pick up the apple, some process other objects, and some incorrectly go to a different room.
$R_{\text{3SPO}}(s_7, s_8^{(i)})$ is computed for each transition.
$R_{\text{novel}}$ is dominant because it indicates whether $s_8^{(i)}$ is a novel state, while $S(s_7) - S(s_8^{(i)})$ remains subordinate due to insufficient data (controlled by $\omega(N_{\text{total}}(s_t))$). 
Policy optimization updates $\pi_\theta$ immediately after this step.
On the agent's 10th visit to the same state $s_7$, the accumulated statistics have shifted: several rollouts from $s_7$ have led to success (the apple was picked and placed on the dining table). Consequently, $S(s_7)$ has dropped to $\approx 0.3$, and the adaptive rollout allocates only $n(s_7) = 3$ rollouts. These fewer rollouts lead to a smaller set of next states $s_8^{(i)}$, most of which have also been visited before, diminishing the importance of $R_{\text{novel}}$. $S(s_7) - S(s_8^{(i)})$ becomes the dominant signal, reinforcing the transitions that lead to the lowest-difficulty successors. The complete reasoning behavior is presented in Appendix~\ref{app:reasoning_behavior}.

\section{Conclusion}
\label{sec:conclusion}

We presented 3SPO, a reinforcement learning algorithm for multi-turn LLM agents. The key innovation is the continuous state score mechanism derived from visit counts and success rates during training, enabling step-wise reward computation, adaptive rollout and post-step policy update. Theoretically, we establish that 3SPO optimally achieves $O(\log T)$ regret under a per-state bandit abstraction and provide sample complexity guarantees for optimal action identification, score distinguishability, and filtering stability. Experiments on ALFWorld and WebShop show that 3SPO outperforms GRPO while using comparable resources to achieve more state exploration and faster convergence. In general, every module in 3SPO is indispensable to its overall performance.



\bibliographystyle{plainnat}
\bibliography{references}
\clearpage
\appendix
\appendix

\section{Complete Proofs}
\label{app:proofs}

This appendix contains the full proofs of all results stated in Section~\ref{subsec:theory}.

\subsection{Concentration of Empirical Estimates}
\label{app:proof-concentration}

\begin{proof}
Each sample of action $a$ yields a reward in $\{0, 1\}$. Given $N(a)$ independent samples with mean $p^*(a)$, Hoeffding's inequality for random variables bounded in $[0, 1]$ gives:
\[
\mathbb{P}\!\left( \bigl|\hat{p}(a) - p^*(a)\bigr| > \epsilon \right) \le 2 \exp\!\bigl(-2 N(a) \epsilon^2\bigr).
\]
Setting $\epsilon(a) = \sqrt{\frac{\log i}{N(a)}}$ yields the per-action bound:
\[
\mathbb{P}\!\left( \bigl|\hat{p}(a) - p^*(a)\bigr| > \sqrt{\frac{\log i}{N(a)}} \right) \le 2 i^{-2}.
\]
Taking a union bound over $A = |\mathcal{A}|$ actions:
\[
\mathbb{P}\!\left( \max_{a \in \mathcal{A}} \bigl|\hat{p}(a) - p^*(a)\bigr| > \sqrt{\frac{\log i}{N(a)}} \right) \le A i^{-2}.
\]
Summing over all $i \ge 1$:
\[
\sum_{i=1}^\infty \mathbb{P}(\mathcal{E}_i^c) \le A \sum_{i=1}^\infty i^{-2} = A \cdot \frac{\pi^2}{6} < \infty.
\]
By the Borel--Cantelli lemma, $\mathcal{E}_i$ occurs for all but finitely many $i$, almost surely.
\end{proof}

\subsection{Score Separation}
\label{app:proof-separation}

\begin{proof}
For any pair of actions $a, b$ with $\Delta_{a,b} = p^*(b) - p^*(a) > 0$, the score difference is:
\begin{equation}
S(a) - S(b) = \exp\!\bigl(-\lambda(t) \hat{p}(a)\bigr) - \exp\!\bigl(-\lambda(t) \hat{p}(b)\bigr).
\end{equation}

By the mean value theorem applied to $f(x) = \exp(-\lambda(t) x)$, there exists $r$ between $\hat{p}(a)$ and $\hat{p}(b)$ such that:
\begin{equation}
S(a) - S(b) = f'(r) \cdot \bigl(\hat{p}(b) - \hat{p}(a)\bigr) = \lambda(t) \exp\!\bigl(-\lambda(t) r\bigr) \cdot \bigl(\hat{p}(b) - \hat{p}(a)\bigr).
\end{equation}

On the good event $\mathcal{E}_i$, the empirical estimates satisfy:
\begin{align}
\hat{p}(a) &\le p^*(a) + \epsilon(a), \\
\hat{p}(b) &\ge p^*(b) - \epsilon(b),
\end{align}
where $\epsilon(a') = \sqrt{\frac{\log i}{N(a')}}$.

For the empirical gap:
\begin{align}
\hat{p}(b) - \hat{p}(a) &\ge p^*(b) - p^*(a) - \epsilon(b) - \epsilon(a) \\
&= \Delta_{a,b} - \epsilon(b) - \epsilon(a).
\end{align}

With $N(a), N(b) \ge T_a = \frac{16 \log i}{\Delta_{a,b}^2}$:

\begin{equation}
\epsilon(a) \le  \frac{\Delta_{a,b}}{4}.
\end{equation}
Similarly $\epsilon(b) \le \Delta_{a,b} / 4$. Therefore:
\begin{equation}
\hat{p}(b) - \hat{p}(a) \ge \Delta_{a,b} - \frac{\Delta_{a,b}}{4} - \frac{\Delta_{a,b}}{4} = \frac{\Delta_{a,b}}{2}.
\end{equation}

For the derivative term: since $0 \le \hat{p}(a) \le r \le \hat{p}(b) \le 1$ and $\lambda(t) \ge 0$, the function $x \mapsto \lambda(t) e^{-\lambda(t) x}$ is decreasing in $x$, so its minimum on $[0, 1]$ is attained at $r = 1$:
\begin{equation}
\lambda(t) \exp\!\bigl(-\lambda(t) r\bigr) \ge \lambda(t) \exp\!\bigl(-\lambda(t)\bigr).
\end{equation}

Combining:
\begin{equation}
S(a) - S(b) \ge \lambda(t) \exp\!\bigl(-\lambda(t)\bigr) \cdot \frac{\Delta_{a,b}}{2}. \quad \qedhere
\end{equation}
\end{proof}

\subsection{Non-Worst Action Allocation Bound}
\label{app:proof-non-worst}

\begin{proof}
Let $a_w = \arg\min_a p^*(a)$ and consider any action $a \neq a_w$. Define $\Delta_a = p^*(a) - p^*(a_w) > 0$.

Following the same derivation as in Appendix~\ref{app:proof-separation}, under the good event $\mathcal{E}_i$ with $N(a), N(a_w) \ge T_a$, the empirical gap satisfies $\hat{p}(a) - \hat{p}(a_w) \ge \frac{\Delta_a}{2}$.

The rollout allocation is $n_i(a') = G_{\max} \cdot S(a') = G_{\max} \cdot \exp(-\lambda(t) \hat{p}(a'))$ for each action $a'$ (omitting the ceiling function for $\lceil x \rceil = x + O(1)$). The ratio of scores for any non-worst action $a$ relative to $a_w$ is:
\begin{equation}
\frac{S(a)}{S(a_w)} = \exp\!\bigl(-\lambda(t) (\hat{p}(a) - \hat{p}(a_w))\bigr) \le \exp\!\left(-\lambda(t) \cdot \frac{\Delta_a}{2}\right).
\end{equation}

Since $\sum_{a'} n_i(a') \ge S(a_w) G_{\max}$:
\begin{equation}
\frac{n_i(a)}{\sum_{a'} n_i(a')} \le \frac{S(a)}{S(a_w)} \le \exp\!\left(-\lambda(t) \cdot \frac{\Delta_a}{2}\right).
\end{equation}

Substituting $\lambda(t) = \alpha \log t$:
\begin{equation}
\frac{n_i(a)}{\sum_{a'} n_i(a')} \le \exp\!\left(-\alpha \log t \cdot \frac{\Delta_a}{2}\right) = t^{-\frac{\alpha \Delta_a}{2}}. \quad \qedhere
\end{equation}
\end{proof}

\subsection{Allocation Regret Bound}
\label{app:proof-regret}

\begin{proof}
The allocation regret measures the loss from allocating rollouts to actions other than $a_w$ (the highest-score, lowest-success-rate action). Formally:
\begin{equation}
R(I) = \mathbb{E}\!\left[ \sum_{i=1}^I \sum_{a \neq a_w} n_i(a) \cdot \Delta_a \right],
\end{equation}
where $n_i(a)$ is the number of rollouts allocated to action $a$ at iteration $i$, and $\Delta_a = p^*(a) - p^*(a_w) \ge 0$ is the gap.

By linearity of expectation, we decompose the regret per action. Let $M_I(a) = \sum_{i=1}^I n_i(a)$ denote the total rollouts to action $a$ over $I$ iterations. Then $R(I) = \sum_{a \neq a_w} \Delta_a \cdot \mathbb{E}[M_I(a)]$.

For each action $a \neq a_w$, define the exploration threshold $T_a = \frac{16 \log I}{\Delta_a^2}$. We split $M_I(a)$ into two phases:
\begin{equation}
M_I(a) = \underbrace{\sum_{i=1}^I n_i(a) \cdot \mathbf{1}_{\{N(a) < T_a\}}}_{\text{Phase 1: exploration}} + \underbrace{\sum_{i=1}^I n_i(a) \cdot \mathbf{1}_{\{N(a) \ge T_a\}}}_{\text{Phase 2: exploitation}}.
\end{equation}

\textbf{Phase 1: Exploration.} Since $n_i(a) \le G_{\max}$ and action $a$ is sampled at most $T_a$ times before score separation is established:
\begin{equation}
\mathbb{E}\!\left[\sum_{i=1}^I n_i(a) \cdot \mathbf{1}_{\{N(a) < T_a\}}\right] \le G_{\max} \cdot T_a = G_{\max} \cdot \frac{16 \log I}{\Delta_a^2}.
\end{equation}

\textbf{Phase 2: Exploitation.} For iterations where $N(a) \ge T_a$, Appendix~\ref{app:proof-non-worst} gives the rollout fraction bound:
\begin{equation}
\frac{n_i(a)}{\sum_{a'} n_i(a')} \le \exp\!\left(-\alpha \log t \cdot \frac{\Delta_a}{2}\right),
\end{equation}
which we denote as $\rho_i(a)$. Since each $n_i(a') \le G_{\max}$, we have $\sum_{a'} n_i(a') \le A \cdot G_{\max}$, so $n_i(a) \le A \cdot G_{\max} \cdot \rho_i(a)$ on the good event $\mathcal{E}_i$. On the bad event $\mathcal{E}_i^c$, we use the trivial bound $n_i(a) \le G_{\max}$ per action. We decompose the expectation:
\begin{equation}
\mathbb{E}\!\left[\sum_{i=1}^I n_i(a) \cdot \mathbf{1}_{\{N(a) \ge T_a\}}\right] \le \underbrace{\sum_{i=1}^I A \, G_{\max} \, \rho_i(a)}_{\mathcal{E}_i \text{ holds}} + \underbrace{\sum_{i=1}^I G_{\max} \cdot \mathbb{P}(\mathcal{E}_i^c)}_{\mathcal{E}_i \text{ fails}}.
\end{equation}

On the good event, since $\lambda(t) = \alpha \log t$ is constant for a fixed trajectory depth $t$, the bound $\rho_i(a) = \exp(-\alpha \log t \cdot \Delta_a/2) = t^{-\alpha \Delta_a/2}$ holds uniformly across all iterations. For $t \ge 2$ and $\alpha \ge \frac{2 \log I}{\Delta_{\min} \log 2}$, we have $\rho_i(a) \le I^{-1}$, so $\sum_{i=1}^I A \, G_{\max} \, \rho_i(a) \le A \, G_{\max} \sum_{i=1}^I I^{-1} = A \, G_{\max} = O(1)$. On the bad event, Appendix~\ref{app:proof-concentration} gives $\mathbb{P}(\mathcal{E}_i^c) \le A i^{-2}$, so the second term is at most $G_{\max} A \sum_{i=1}^\infty i^{-2} = G_{\max} A \cdot \pi^2/6 = O(1)$.

Summing over all $a \neq a_w$:
\begin{equation}
R(I) \le \sum_{a \neq a_w} \Delta_a \left(G_{\max} \cdot \frac{16 \log I}{\Delta_a^2} + O(1)\right) = \sum_{a : \Delta_a > 0} \frac{16 G_{\max} \log I}{\Delta_a} + O(A) = O(\log I). \quad \qedhere
\end{equation}

This logarithmic rate shows that 3SPO incurs only $O(\log I)$ allocation overhead relative to an oracle unresolved-state sampler under the per-state bandit abstraction.
\end{proof}

\section{Algorithm Pseudocode}
\label{app:algorithm}

\begin{algorithm}[H]
\DontPrintSemicolon
\SetAlgoLined
\KwIn{Policy $\pi_\theta$, task set $\mathcal{D}$, $G_{\max}$, max step $T_\text{max}$}
\KwOut{Optimized policy $\pi_\theta^*$}
Initialize $N_{\text{total}}(s), N_{\text{success}}(s) \gets 0,\ S(s) \gets 1$ for all $s$\;
\ForEach{epoch $e$}{
    \ForEach{task $x \in \mathcal{D}$}{
        Observe $s_0$, $\text{frontier} \gets [\,]$, $cur \gets 0$\;
        \While{true}{
            Update $N_{\text{total}}(s_\text{cur}),\ N_{\text{success}}(s_\text{cur})$\;
            Update $S(s_{\text{cur}})$ via Eq.~\eqref{eq:state_score}\;
            Generate $n = \lceil G_{\max} \cdot S(s_\text{cur}) \rceil$ rollouts from $s_{\text{cur}}$\;
            Compute $R_i = R_{\text{3SPO}}(s_{\text{cur}}, s_{\text{next}, i})$ for each rollout $i \in I_\text{cur}$\;
            Rank: $I_\text{cur} \gets \text{argsort}(R)$ descending\;
            Update $\theta$ via Eq.~\eqref{eq:objective}\;
            \uIf{$I_\text{cur} = \emptyset$ or $S(s_{\text{cur}}) = 0$}{
                \eIf{$\text{frontier} = [\,]$}{
                    break\;
                }{
                    $cur \gets \text{frontier.pop()}$\;
                }
            }
            \Else{
                $i^* \gets I_\text{cur}.\text{pop}()$\;
                \If{$|I_\text{cur}| > 0$}{
                    $\text{frontier.push}(I_\text{cur})$\;
                }
                $cur \gets cur + 1$\;
                \eIf{$cur \ge T_\text{max}$ or $s_{\text{next}, i^*}$ is terminal}{
                    \eIf{$\text{frontier} = [\,]$}{
                        break\;
                    }{
                        $cur \gets \text{frontier.pop()}$\;
                    }
                }{
                    $s_{\text{cur}} \gets s_{\text{next}, i^*}$\;
                }
            }
        }
    }
}
\caption{State-Score-Supervised Policy Optimization (3SPO)}
\label{alg:3spo}
\end{algorithm}
Algorithm~\ref{alg:3spo} outlines the full 3SPO training procedure. For each task, the agent maintains a frontier stack that stores branching points encountered during trajectory execution. At each state $s_{\text{cur}}$, the algorithm first updates the visit and success counters, then computes the state score $S(s_{\text{cur}})$ via Eq.~\eqref{eq:state_score}. The number of rollouts $n = \lceil G_{\max} \cdot S(s_{\text{cur}}) \rceil$ is allocated adaptively: high-score (unresolved) states generate more rollouts, while low-score (mastered or intractable) states generate fewer or none. The resulting rollouts are ranked by their 3SPO reward $R_{\text{3SPO}}$, and the policy parameters $\theta$ are updated using the ranked objective in Eq.~\eqref{eq:objective}. The best-ranked transition determines the next state; all other ranked transitions are pushed onto the frontier for later exploration. When no valid transitions remain at the current state or the score drops to zero, the agent backtracks by popping from the frontier, enabling systematic exploration of alternative branches. The episode terminates when both the current state has no successors and the frontier is empty.
\section{Experiment Details}
\label{app:experiments}

\subsection{Hyperparameters}
\label{app:hyperparams}

\begin{table}[H]
\centering
\caption{Hyperparameters for 3SPO and baselines.}
\label{tab:hyperparams}
\begin{tabular}{l|c}
\toprule
\textbf{Parameter} & \textbf{Value} \\
\midrule
Learning rate & $1\times 10^{-6}$ \\
Group size & $8$ \\
Max rollouts per state $G_{\max}$ & $8$ \\
Max prompt length (ALFWorld) & $2048$ \\
Max prompt length (WebShop) & $4096$ \\
Max response length & $512$ \\
Max steps per episode (ALFWorld) & $50$ \\
Max steps per episode (WebShop) & $30$ \\
KL coefficient $\beta_{\text{KL}}$ & $0.01$ \\
Invalid action penalty coef & $0.1$ \\
Discount factor $\gamma$ & $0.95$ \\
vLLM GPU memory util. & $0.6$ \\
Temperature (eval) & $0.4$ \\
Total epochs (GRPO/PPO/GiGPO/HGPO/3SPO) & $400$ \\
\midrule
HGPO: mode (ALFWorld) & \texttt{mean\_std\_norm} \\
HGPO: mode (WebShop) & \texttt{mean\_norm} \\
HGPO: weight\_type & \texttt{length} \\
HGPO: $\alpha_{\text{length}}$ & $1.0$ \\
HGPO: base\_group & \texttt{False} \\
3SPO: $\alpha$ & $50$ \\
3SPO: $\xi$ (failure threshold) & $10$ \\
3SPO: $\zeta$ (success threshold) & $0.1$ \\
3SPO: $\omega_k$ (weight decay rate) & $0.1$ \\
\bottomrule
\end{tabular}
\end{table}

\subsection{Environments}
\label{app:env}

\textbf{ALFWorld.} We use the TextWorld-inspired subset with 3,553 training tasks and 274 validation tasks (140 in-distribution \textit{valid\_seen} and 134 out-of-distribution \textit{valid\_unseen}). Each task involves a goal such as ``put a hot apple on the dining table.'' The agent receives a textual observation and must produce actions until task completion or episode termination (50 steps max).

\textbf{WebShop.} We use the full WebShop benchmark with 12,087 human-annotated shopping instructions. Each task provides a user query such as ``I want a leather wallet under 30 dollars with free shipping.'' We evaluate on the first 500 instructions and train on the remaining.

\subsection{Prompt Templates}
\label{app:prompts}

All agents use a consistent ReAct-style prompt format across both environments, instructing the model to reason within \texttt{<think>\ldots</think>} tags and output actions within \texttt{\textless action\textgreater\ldots\textless/action\textgreater} tags. The prompt templates are defined in the \texttt{verl-agent} codebase and include the task description, recent interaction history, current observation, and admissible actions.

\textbf{ALFWorld.} The prompt includes the task description, recent interaction history, current observation, and admissible actions:

\begin{verbatim}
You are an expert agent operating in the ALFRED Embodied
Environment. Your task is to: {task_description}
Prior to this step, you have already taken {step_count}
step(s). Below are the most recent {history_length}
observations and the corresponding actions you took:
{action_history}
You are now at step {current_step} and your current
observation is: {current_observation}
Your admissible actions of the current situation are:
[{admissible_actions}].

Now it's your turn to take an action.
You should first reason step-by-step about the current
situation. This reasoning process MUST be enclosed within
<think> </think> tags. Once you've finished your
reasoning, you should choose an admissible action for
current step and present it within <action> </action> tags.
\end{verbatim}

\textbf{WebShop.} The prompt follows the same structure with environment-specific descriptions:

\begin{verbatim}
You are an expert autonomous agent operating in the
WebShop e-commerce environment.
Your task is to: {task_description}.
Prior to this step, you have already taken {step_count}
step(s). Below are the most recent {history_length}
observations and the corresponding actions you took:
{action_history}
You are now at step {current_step} and your current
observation is: {current_observation}.
Your admissible actions of the current situation are:
[{available_actions}].

Now it's your turn to take one action for the current step.
You should first reason step-by-step about the current
situation, then think carefully which admissible action
best advances the shopping goal. This reasoning process
MUST be enclosed within <think> </think> tags.
Once you've finished your reasoning, you should choose an
admissible action for current step and present it within
<action> </action> tags.
\end{verbatim}

\subsection{Sensitivity Analysis}
\label{app:sensitivity}

\paragraph{Score thresholds $\xi$ and $\zeta$.}
The state score function $S(s_t)$ in Eq.~\eqref{eq:state_score} includes an indicator gate controlled by two thresholds: the failure count threshold $\xi$ and the success rate threshold $\zeta$. These thresholds determine which states are considered ``learnable'' --- states with $N_{\text{fail}}(s_t) \ge \xi$ (too many accumulated failures) and $N_{\text{success}}/N_{\text{total}} \le \zeta$ (success rate too low) are filtered out with zero score, preventing the algorithm from wasting computation on intractable states.

\begin{figure}[H]
\centering
\begin{minipage}{0.48\linewidth}
\centering
\captionof{table}{Sensitivity of $\xi$ and $\zeta$.}
\label{tab:thresholds}
\begin{tabular}{c|c|c}
\toprule
$\xi$ & $\zeta$ & Suc. Rate \\
\midrule
5 & 0.05 & 90.1 \\
10 & 0.1 & \textbf{95.4} \\
20 & 0.2 & 93.3 \\
50 & 0.3 & 92.1 \\
\bottomrule
\end{tabular}
\end{minipage}
\hfill
\begin{minipage}{0.48\linewidth}
\centering
\captionof{table}{Sensitivity of $\alpha$.}
\label{tab:alpha}
\begin{tabular}{c|cc}
\toprule
$\alpha$ & In-Suc. & Out-Suc. \\
\midrule
35 & 93.0 & 86.9 \\
50 & \textbf{95.4} & \textbf{93.2} \\
70 & 91.8 & 86.3 \\
\bottomrule
\end{tabular}
\end{minipage}
\end{figure}

Table~\ref{tab:thresholds} shows the results. The default configuration ($\xi=10$, $\zeta=0.1$) performs best. When thresholds are too restrictive ($\xi=5$, $\zeta=0.05$), states are filtered out prematurely --- some states with moderate learning potential are incorrectly classified as intractable, losing $-5.3\%$ performance. When thresholds are too permissive ($\xi=50$, $\zeta=0.3$), the gate fails to prune genuinely intractable states, and the algorithm wastes rollout budget on them, losing $-3.3\%$. The moderate values ($\xi=10$, $\zeta=0.1$) strike a balance: states that have failed 10+ times with success rate below 10\% are unlikely to yield meaningful progress even with additional rollouts, so truncating them frees computation for more learnable states.

\paragraph{Annealing rate $\alpha$.}
The annealing rate $\alpha$ controls the depth-dependent factor $\lambda(t) = \alpha \log t$ in the state score, governing how aggressively the score discriminates between actions at different trajectory depths. Larger $\alpha$ amplifies the exponential gap between high- and low-success-rate actions, sharpening the rollout allocation.

Table~\ref{tab:alpha} shows the results. When $\alpha$ is too small (35), the score contrast between actions is weak --- the ratio $S(a)/S(a_w) = \exp(-\alpha \log t \cdot \Delta_a/2)$ is too close to 1 for early trajectory steps, leading to nearly uniform rollout allocation and losing $-2.4\%$ on In-Success. When $\alpha$ is too large (70), the allocation becomes overly aggressive: actions with slightly lower success rates are almost entirely starved of rollouts even when they still have learning potential, losing $-3.6\%$. The default ($\alpha=50$) provides the best balance between exploration depth and allocation discrimination.

\subsection{Compute Resources}
\label{app:compute_resources}

All experiments were conducted on a server equipped with 4 NVIDIA H200 GPUs. Table~\ref{tab:compute} summarizes the per-experiment and total compute requirements.

\begin{table}[H]
\centering
\caption{Compute resources and experiment runtime.}
\label{tab:compute}
\begin{tabular}{l|c|c}
\toprule
\textbf{Configuration} & \multicolumn{2}{c}{\textbf{Value}} \\
\midrule
GPU & \multicolumn{2}{c}{4$\times$ NVIDIA H200} \\
Peak GPU memory usage & \multicolumn{2}{c}{$\sim$564 GB total (141 GB/GPU)} \\
CPU & \multicolumn{2}{c}{64-core AMD EPYC} \\
System memory & \multicolumn{2}{c}{256 GB DDR5} \\
\midrule
\textbf{Environment} & \textbf{Epochs} & \textbf{Runtime (per method)} \\
\midrule
ALFWorld (Qwen2.5-1.5B) & 400 & $\sim$6 days \\
WebShop (Qwen2.5-1.5B) & 400 & $\sim$6 days \\
\midrule
Total (5 methods $\times$ 2 envs) & -- & $\sim$60 GPU-days \\
\bottomrule
\end{tabular}
\end{table}

Each experimental run trains one method (3SPO, GRPO, PPO, HGPO, or GiGPO) on one environment (ALFWorld or WebShop) from scratch to convergence. The total reported compute covers all main results (Table~\ref{tab:main_results}), ablation studies (Table~\ref{tab:ablation}), and efficiency comparisons (Table~\ref{tab:efficiency}). Preliminary experiments---including hyperparameter sweeps for $\alpha$, $\xi$, $\zeta$ and architecture ablations---required an additional $\sim$30 GPU-days, which is roughly half of the reported total.

\section{Limitations}
\label{app:limitations}

Despite its strengths, 3SPO has several limitations:

\textbf{Requires exact state matching.} The state score mechanism relies on identifying identical environment states across rollouts. In continuous or high-dimensional state spaces where exact state matching is infeasible, approximate state hashing or embedding-based clustering would be needed, which introduces additional complexity and potential hash collisions.

\textbf{Limited to discrete action spaces.} The current formulation assumes a discrete set of admissible actions at each state. Extending 3SPO to continuous control or large-scale language generation settings would require modifications to the state-score computation and rollout allocation mechanism.

\section{Broader Impact}
\label{app:broader_impact}

3SPO enables more efficient training of LLM agents, reducing the number of environment interactions needed to achieve strong performance. By improving sample efficiency, 3SPO makes it more feasible to train capable agents on commodity hardware, potentially democratizing access to agent-building tools for researchers with limited computational resources. The adaptive rollout mechanism substantially reduces energy consumption compared to uniform sampling strategies, as computation is concentrated on states with genuine learning potential rather than wasted on already-mastered configurations. The step-level credit assignment mechanism also improves the interpretability of agent decision-making: the state score provides a transparent, statistically grounded signal that explains why certain states receive more attention and rollout budget. Beyond efficiency gains, 3SPO contributes to the broader understanding of credit assignment in sequential decision making, offering a principled framework that bridges trajectory-level supervision and fine-grained step-level learning. The improved generalization capability demonstrated on out-of-distribution tasks suggests potential for educational applications, such as training more robust AI tutors that can handle diverse student interaction patterns. Finally, the deterministic nature of state-score computation enhances reproducibility compared to methods relying on learned value functions, as the score depends only on observable interaction statistics rather than stochastic network initialization.

\section{Reasoning Behavior}
\label{app:reasoning_behavior}

We next present the complete trajectory and reasoning behavior of multi-turn interaction between agent and ALFWorld as discussed in Sec~\ref{subsec:case_study}.

\setcounter{stepnum}{1}
\begin{envbox}
-= Welcome to TextWorld, ALFRED! =-\\
Your task is to: put the dirty apple on the dining table. You are in the middle of a room. Looking quickly around you, you see a cabinet 9, a cabinet 8, a cabinet 7, a cabinet 6, a cabinet 5, a cabinet 4, a cabinet 3, a cabinet 2, a cabinet 1, a coffeemachine 1, a countertop 2, a countertop 1, a drawer 13, a drawer 12, a drawer 11, a drawer 10, a drawer 9, a drawer 8, a drawer 7, a drawer 6, a drawer 5, a drawer 4, a drawer 3, a drawer 2, a drawer 1, a fridge 1, a garbagecan 1, a microwave 1, a sinkbasin 1, a stoveburner 6, a stoveburner 5, a stoveburner 4, a stoveburner 3, a stoveburner 2, a stoveburner 1, a toaster 1, and a diningtable 1.
\end{envbox}

\begin{agentbox}
\texttt{} <think>To put the dirty apple on the dining table, I need to first locate the apple. Apples are commonly stored in cabinets or countertops. I will check the cabinets first, as they are a common place to keep fruits.</think> \texttt{}\\
\texttt{<action>}go to cabinet 1\texttt{</action>}
\end{agentbox}
\stepcounter{stepnum}

\begin{envbox}
You arrive at cabinet 1. The cabinet 1 is closed.
\end{envbox}

\begin{agentbox}
\texttt{}
<think>\\
1. The task is to put the dirty apple on the dining table.\\
2. I am currently at cabinet 1, and the cabinet is closed. This means I need to open the cabinet to check if there is a dirty apple inside.\\
3. If there is a dirty apple in the cabinet, I will take it out and then go to the dining table to place it.\\
4. If there is no dirty apple in the cabinet, I will need to check other locations such as other cabinets, the counter top, or the sink basin, but cabinet 1 is the most likely place to find the apple right now.\\
</think>
\texttt{}\\
\texttt{<action>}open cabinet 1\texttt{</action>}
\end{agentbox}
\stepcounter{stepnum}

\begin{envbox}
You open the cabinet 1. The cabinet 1 is open. In it, you see nothing.
\end{envbox}

\begin{agentbox}
\texttt{} <think>\\
1. The task is to put the dirty apple on the dining table.\\
2. I opened cabinet 1 and found it empty. There is no apple here.\\
3. I need to check the next cabinet, cabinet 2, to continue searching for the apple.\\
</think> \texttt{}\\
\texttt{<action>}go to cabinet 2\texttt{</action>}
\end{agentbox}
\stepcounter{stepnum}

\begin{envbox}
You arrive at cabinet 2. The cabinet 2 is closed.
\end{envbox}

\begin{agentbox}
\texttt{}
<think>\\
1. The task is to put the dirty apple on the dining table.\\
2. I am currently at cabinet 2, and the cabinet is closed. This means I need to open the cabinet to check if there is a dirty apple inside.\\
3. If there is a dirty apple in the cabinet, I will take it out and then go to the dining table to place it.\\
4. If there is no dirty apple in the cabinet, I will need to check other locations such as other cabinets, the counter top, or the sink basin, but cabinet 2 is the most likely place to find the apple right now.\\
</think>
\texttt{}\\
\texttt{<action>}open cabinet 2\texttt{</action>}
\end{agentbox}
\stepcounter{stepnum}

\begin{envbox}
You open the cabinet 2. The cabinet 2 is open. In it, you see nothing.
\end{envbox}

\begin{agentbox}
\texttt{} 
<think>\\
1. The task is to put the dirty apple on the dining table.
2. I opened cabinet 1 and 2 and found nothing. I need to continue searching other cabinets.
3. I will go to cabinet 3 next to check for the apple.
\\</think> \texttt{}\\
\texttt{<action>}go to cabinet 3\texttt{</action>}
\end{agentbox}
\stepcounter{stepnum}

\begin{envbox}
You arrive at cabinet 3. The cabinet 3 is closed.
\end{envbox}

\begin{agentbox}
\texttt{}
<think>\\
1. The task is to put the dirty apple on the dining table.\\
2. I am currently at cabinet 3, and the cabinet is closed. This means I need to open the cabinet to check if there is a dirty apple inside.\\
3. If there is a dirty apple in the cabinet, I will take it out and then go to the dining table to place it.\\
4. If there is no dirty apple in the cabinet, I will need to check other locations such as other cabinets, the counter top, or the sink basin, but cabinet 3 is the most likely place to find the apple right now.\\
</think>
\texttt{}\\
\texttt{<action>}open cabinet 3\texttt{</action>}
\end{agentbox}
\stepcounter{stepnum}

\begin{envbox}
You open the cabinet 3. The cabinet 3 is open. In it, you see a dirty apple 1 and a fork 1.
\end{envbox}

\begin{agentbox}
\texttt{}
<think>\\
1. The task is to put the dirty apple on the dining table.\\
2. I have just opened the cabinet 3 and found a dirty apple 1 inside. This is the item I need.\\
3. Since I have found the dirty apple, my next step is to pick it up from the cabinet.\\
4. After picking up the apple, I will need to go to the dining table to place it.\\
</think>
\texttt{}\\
\texttt{<action>}pick up dirty apple 1 from cabinet 3\texttt{</action>}
\end{agentbox}
\stepcounter{stepnum}

\begin{envbox}
You pick up the dirty apple 1 from the cabinet 3.
\end{envbox}

\begin{agentbox}
\texttt{}
<think>\\
1. The task is to put the dirty apple on the dining table.\\
2. I have just picked up the dirty apple 1 from cabinet 3.\\
3. My next step is to go to the dining table to place the apple.\\
4. The dining table is dining table 1 in the environment, so I will head there now.\\
</think>
\texttt{}\\
\texttt{<action>}go to dining table 1\texttt{</action>}
\end{agentbox}
\stepcounter{stepnum}

\begin{envbox}
You arrive at dining table 1.
\end{envbox}

\begin{agentbox}
\texttt{}
<think>\\
1. The task is to put the dirty apple on the dining table.\\
2. I am now at dining table 1, holding the dirty apple 1.\\
3. My next step is to place the dirty apple 1 on the dining table to complete the task.\\
</think>
\texttt{}\\
\texttt{<action>}place dirty apple 1 on dining table 1\texttt{</action>}
\end{agentbox}
\stepcounter{stepnum}

\begin{envbox}
You place the dirty apple 1 on the dining table 1. The task is completed.
\end{envbox}


\end{document}